\documentclass[conference]{IEEEtran}
\usepackage{times}

\usepackage[numbers]{natbib}
\usepackage{multicol}
\usepackage[bookmarks=true]{hyperref}

\IEEEoverridecommandlockouts

\usepackage[ruled,vlined]{algorithm2e} 
\usepackage[utf8]{inputenc} 
\usepackage[T1]{fontenc}    
\usepackage{amsfonts}       
\usepackage{xcolor}         
\usepackage{amsmath, amssymb, bm, xargs, booktabs}
\usepackage{url}

\usepackage{graphicx}
\usepackage{etoolbox}

\begin{document}

\title{Diffusion-based 
Inverse
Observation Model for Artificial Skin}

\author{
\IEEEauthorblockN{Ante Mari\'c\IEEEauthorrefmark{1}\IEEEauthorrefmark{2},
Julius Jankowski\IEEEauthorrefmark{1}\IEEEauthorrefmark{2},
Giammarco Caroleo\IEEEauthorrefmark{3},
Alessandro Albini\IEEEauthorrefmark{3},
Perla Maiolino\IEEEauthorrefmark{3},
Sylvain Calinon\IEEEauthorrefmark{1}\IEEEauthorrefmark{2}}
\IEEEauthorblockA{\IEEEauthorrefmark{1} Idiap Research Institute, Martigny, Switzerland}
\IEEEauthorblockA{\IEEEauthorrefmark{2} École Polytechnique Fédérale de Lausanne (EPFL), Switzerland}
\IEEEauthorblockA{\IEEEauthorrefmark{3} Oxford Robotics Institute (ORI), University of
Oxford, UK}%
\thanks{Corresponding author: ante.maric@idiap.ch}%
}

\maketitle


\begin{abstract}
Contact-based estimation of object pose is challenging due to discontinuities and ambiguous observations that can correspond to multiple possible system states. This multimodality makes it difficult to efficiently sample valid hypotheses while respecting contact constraints. Diffusion models can learn to generate samples from such multimodal probability distributions through denoising algorithms. We leverage these probabilistic modeling capabilities to learn an inverse observation model conditioned on tactile measurements acquired from a distributed artificial skin. We present simulated experiments demonstrating efficient sampling of contact hypotheses for object pose estimation through touch.
\end{abstract}

\IEEEpeerreviewmaketitle


\section{Introduction}
Estimating object pose through contact is difficult because of partial observability caused by discontinuous observations that can correspond to multiple candidate poses. Generating pose hypotheses for a given contact-induced observation therefore requires sampling multimodal probability distributions subject to contact constraints.
Earlier work relies on the use of F/T observations and random sampling of pose hypotheses within the vicinity of prior estimates in Bayesian estimation frameworks~\cite{petrovskayaGlobalLocalizationObjects2011}.
Subsequent methods aim to reduce the number of physically inconsistent samples by relying on contact constraints and tactile sensing. For example,~\cite{kovalPoseEstimationPlanar2015} samples a manifold of valid contact configurations in a particle filtering framework with tactile observations.
Recent approaches attempt to integrate such observations and constraints in a scalable manner by using learning-based methods to sample pose hypotheses~\cite{rostelLearningStateEstimator2022} or to learn observation models of visuotactile sensors~\cite{sodhiLEOLearningEnergybased2022, bauza2023tac} for object localization with sensorized grippers.

Compared to smaller-scale manipulation tasks considered in these works, we target whole-body manipulation settings that may involve clutter and intermittent contact. We aim to leverage distributed tactile observations of an artificial skin~\cite{maiolino2013flexible}, which can provide feedback from various parts of the robot body that is beneficial in these scenarios~\cite{caroleo2024proxy}. 
To sample object pose hypotheses based on tactile observations provided by the artificial skin, we learn an inverse tactile observation model using a denoising diffusion probabilistic model (DDPM)~\cite{ho2020ddpm}, which has been shown to provide strong generative capabilities in pose estimation under partial observability~\cite{xu20246d, moller2024particlebased6dobjectpose}.
We evaluate the sampling efficiency improvements enabled by such tactile conditioning in simulated planar pose estimation experiments using particle filtering, with the artificial skin mounted on a cylindrical end-effector.
\begin{figure}[t]
  \centering
  \includegraphics[width=\linewidth]{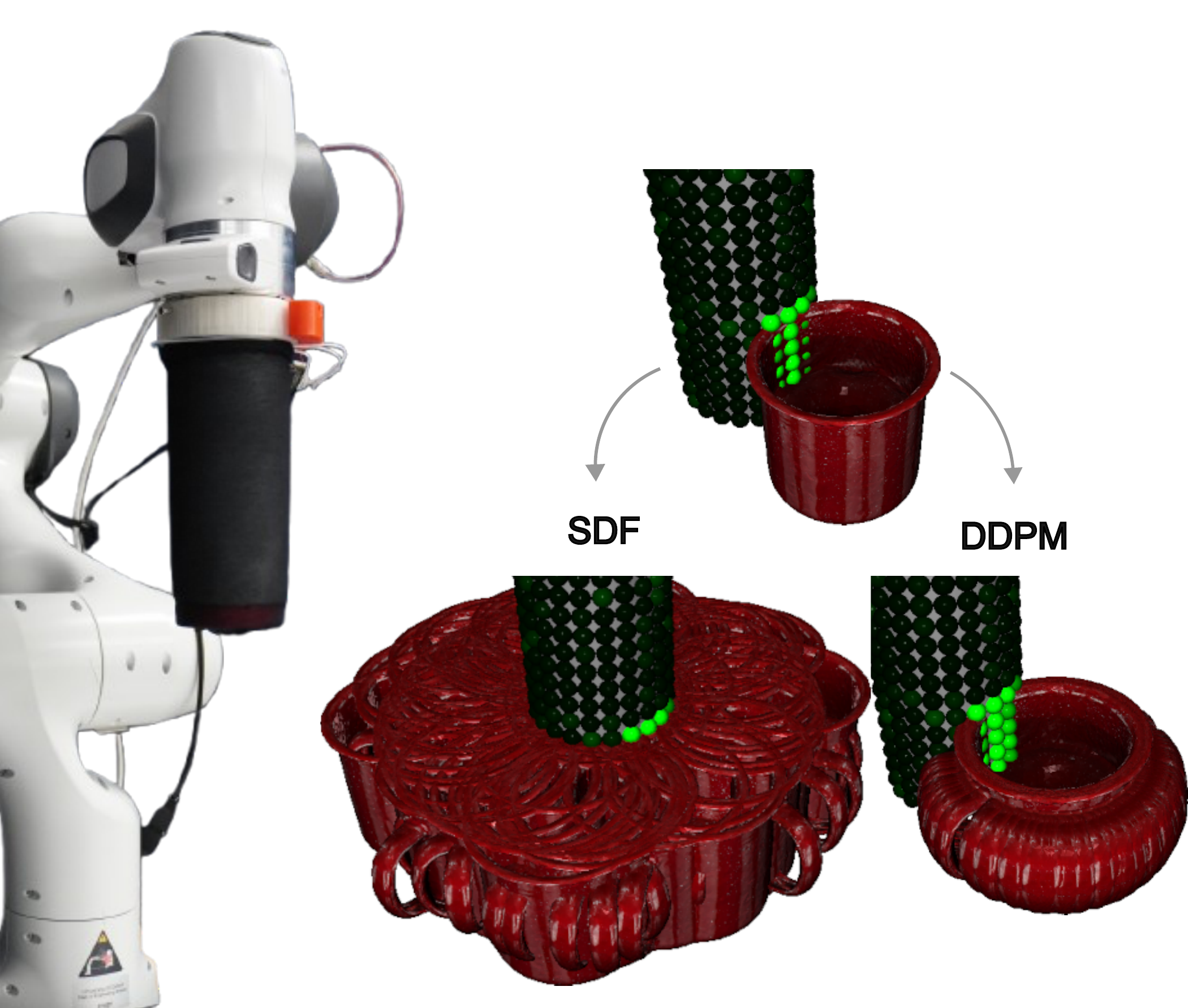}
  \caption{\textit{(Left)} Targeted setup of a 7-axis manipulator with \textit{CySkin} mounted on a cylindrical end-effector. \textit{(Right)} Simulated experiment: $100$ contact configurations generated for \texttt{025\_mug} using SDF projection and a DDPM conditioned on tactile activations. Taxels are represented as spheres, with green color intensity proportional to their activation values.}
  \label{fig:intro}
\end{figure}
\begin{figure}[t]
  \centering
  \includegraphics[width=\linewidth]{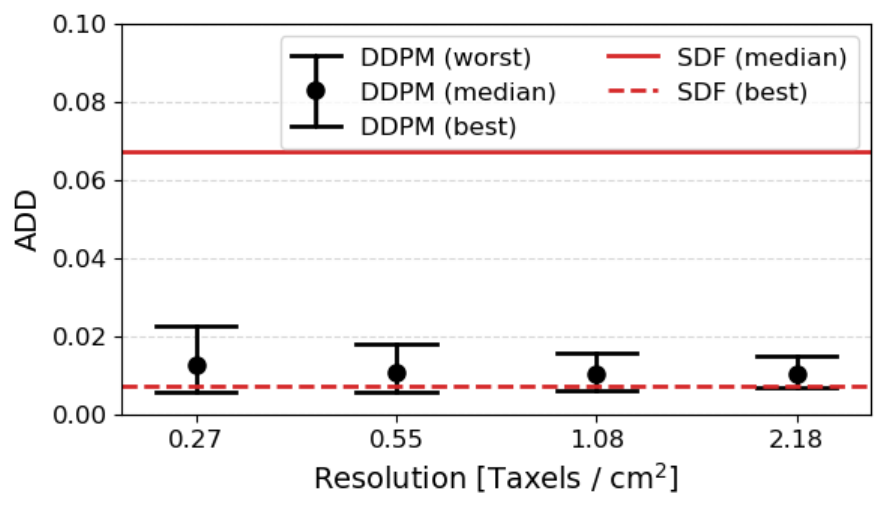}
  \caption{Median, highest, and lowest ADD errors for $100$ particles sampled across $1000$ random contact configurations with objects from the test set\protect\footnotemark. Errors are shown  for different resolutions of the tactile array.}
  \label{fig:comp}
\end{figure}

\section{Method}
\label{sec:method}
We assume that an observation $\bm{z}^\mathrm{tax}$ acquired from the artificial skin is a sample from a conditional probability distribution over taxel activations
\begin{equation}
\label{eq:obs_probs}
    \bm{z}^\mathrm{tax} \sim p\left(\cdot | \bm{q}^o \right),
\end{equation}
where the underlying distribution is described by a nominal observation model with continuous taxel activations, and $\bm{q}^o=\left(x^o, y^o,\cos(\theta),\sin(\theta)\right)$ represents planar object pose  expressed in the sensor frame.
We wish to sample the \textit{inverse} of this observation model by considering a distribution over object poses for a given observation
\begin{equation}
\label{eq:proposal}
    \bm{q}^o \sim p(\cdot | \bm{z}^\mathrm{tax}).
\end{equation}

For an object of known shape, we generate a dataset $D = \{ {\bm{q}}^{o}_n, \bm{z}^\mathrm{tax}_{n} \}_{n=1}^{N_{D}}$ of simulated state-observation pairs where the object is in contact with the sensor.
Contact configurations are synthesized using single-step projection of $N_D=10^4$ uniformly sampled initial object poses $\bm{q}^{o}_{\text{init}, n}$ using a signed distance field (SDF) $\sigma$ and its gradient. Namely, we apply a translation $\bm{t}^o$ to project each random sample into contact with the artificial skin
\begin{align}
    \bm{t}_n^o &=  - \left(\sigma(\bm{q}^{o}_{\text{init}, n}) + \delta\right)\frac{\nabla\sigma(\bm{q}^{o}_{\text{init}, n})}{\|\nabla\sigma(\bm{q}^{o}_{\text{init}, n})\|},
\end{align}
$\forall n = 1, \dots, N_D$, where $\delta \sim \mathcal{U}([0, \delta_{\text{max}}])$ is a sampled penetration depth that approximates compliance of the artificial skin. Each contact configuration is paired with a tactile observation $\bm{z}^{\text{tax}}_n$ generated by querying the observation model~\eqref{eq:obs_probs}.%
\footnotetext{Object set: \texttt{002\_master\_chef\_can}, \texttt{003\_cracker\_box}, \texttt{006\_mustard\_bottle}, \texttt{016\_pear}, \texttt{019\_pitcher\_base}, \texttt{021\_bleach\_cleanser}, \texttt{024\_bowl}, \texttt{025\_mug}, \texttt{035\_power\_drill}, \texttt{061\_foam\_brick}.}%

We train a DDPM~\cite{ho2020ddpm} with a 4-layer feedforward NN as the noise predictor $\bm{\epsilon}_\theta$.
The model is trained to reverse a diffusion process over $T=100$ steps conditioned on $\bm{z}^\mathrm{tax}$.
During training, random diffusion steps $t$ are sampled and noise $\bm{\epsilon} \sim \mathcal{N}(0, \bm{I})$ is added to the poses from ground-truth state-observation pairs.
Model parameters $\theta$ are optimized by minimizing the expected MSE loss over the predicted noise $\hat{\bm{\epsilon}} = \bm{\epsilon}_\theta(\bm{\tilde{q}}^o_t, \bm{z}^{\text{tax}}, t)$, with $\bm{\tilde{q}}^o_t$ representing a diffused particle at timestep $t$.
During inference, pose samples are generated by iteratively applying denoising steps for $t = T, \dots, 1$, starting from $\bm{\tilde{q}}^o_T \sim \mathcal{N}(0, \bm{I})$.
This is done in parallel for $S$ pose hypotheses, finally generating a set $\{\tilde{\bm{q}}^o\}_{i=1}^{S}$ that approximates samples from~\eqref{eq:proposal} for a given observation. Figure~\ref{fig:intro} shows qualitative examples of generated pose hypotheses.
\begin{figure}[t]
    \centering
    \vspace{4px}
    \includegraphics[width=\linewidth]{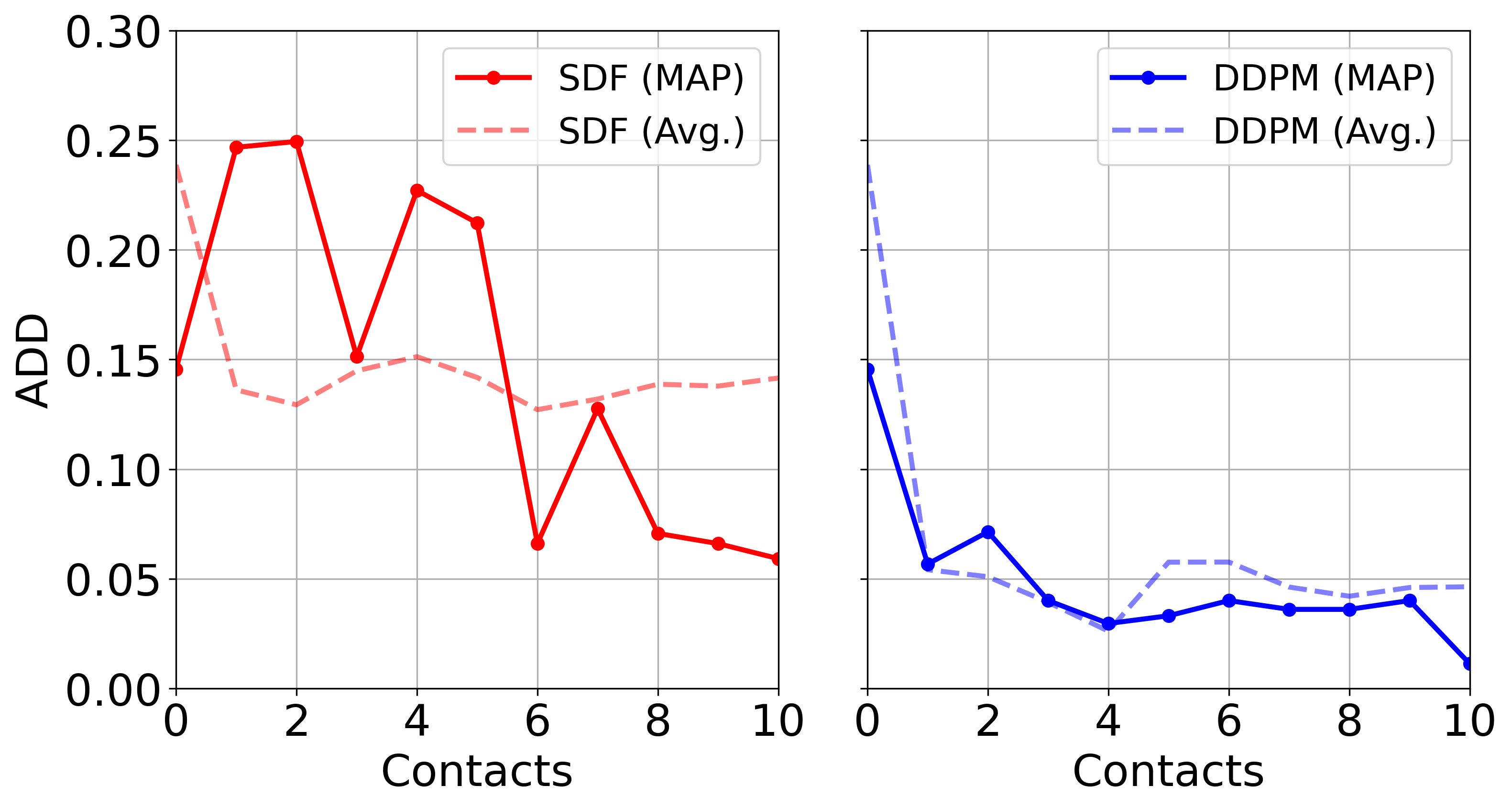}
\caption{Example run showing the ADD error of \textit{maximum a posteriori} (MAP) and weighted average estimates of planar pose for a static box object. \textit{(Left)} using SDF projection for particle proposal. \textit{(Right)} using the \mbox{DDPM-based} inverse observation model for particle proposal. Both cases use the same particle filtering framework with a $100$-particle belief.}
  \label{fig:comp_pf}
\end{figure}

At runtime, we use the generated pose hypotheses for estimation in a particle filtering framework. Namely, samples generated on contact with a static object are used to replace low-likelihood particles as shown in Alg.~\ref{alg:PF}. Periodic resampling and reductions in the number of injected particles are then used to converge toward an estimate.

\section{Simulated Experiments}
\label{sec:result}
We evaluate the improvements gained by conditioning pose hypotheses on tactile observations from the artificial skin by comparing DDPM-generated samples to those generated using SDF projection described in Section~\ref{sec:method}. Figure~\ref{fig:comp} shows a comparison of Average Distance of Model Points (ADD) metrics on objects from the YCB dataset~\cite{calli2017ycb}.
Pose estimation error of the proposed particle filtering framework with respect to the number of contacts with a static box object is shown in Figure~\ref{fig:comp_pf}.
\begin{algorithm}[t]
\caption{Particle Injection on Contact.}
\label{alg:PF}
\SetKwInOut{Input}{Input}
\SetKwInOut{Output}{Output}

\Input{Belief $b = \{(\bm{q}^o_i, w_i)\}_{i=1}^N$, $\bm{z}^\mathrm{tax}$, $S$}
\Output{Updated belief $b'$}

Sort weights $\{w_i\}$ in ascending order\;
Generate $\{\tilde{\bm{q}}_i^o\}_{i=1}^{S}$ with DDPM;\quad\tcp{$\sim p(\cdot \mid \bm{z}^\mathrm{tax})$}\
\For{$i = 1$ \KwTo $S$}{
    Replace $(\bm{q}^o_i, w_i) \gets (\bm{\tilde{q}}^o_i, \bar{w})$, where $\bar{w} = \frac{1}{N} \sum_j w_j$\;
}
Normalize weights: $w_i \gets \frac{w_i}{\sum_j w_j}$\;
\end{algorithm}
The average sampling time for $S=100$ samples is $15~ms$ on an 8-core \textit{Apple M2} chip.

\section{Conclusion}
\label{sec:discussion}
The results of simulated experiments (Fig.~\ref{fig:comp}) demonstrate that learning an inverse observation model by conditioning a DDPM on tactile observations improves the efficiency of pose hypothesis sampling during contact. 
Figure~\ref{fig:comp_pf} also shows that generating contact hypotheses consistent with observations from a distributed artificial skin reduces the number of contacts needed for an accurate pose estimate and leads to faster convergence. Future work will focus on further experimental validation, extensions to pose tracking tasks, and deployment in the real world with artificial skin embedded on different parts of the robot body.

\section*{Acknowledgments}
This work was supported by the 
the European Commission’s Horizon Europe Program through the SESTOSENSO project (\url{http://sestosenso.eu/}) and the INTELLIMAN project (\url{https://intelliman-project.eu/}).

\bibliographystyle{ieeetr}
\bibliography{main.bib}  

\end{document}